\documentclass[english,a4paper,11pt]{article}
\usepackage[top = 2cm, bottom = 2cm,left = 2cm, right = 2cm]{geometry}
\usepackage[T1]{fontenc}
\usepackage[latin9, utf8]{inputenc}
\usepackage{bm}
\usepackage{amsmath}
\usepackage{amssymb}
\usepackage{nccmath}
\usepackage{graphicx}

\usepackage{caption}
\usepackage[lofdepth,lotdepth]{subfig}

\usepackage[linesnumbered,ruled,vlined]{algorithm2e}
\usepackage{mwe}
\usepackage{babel}
\usepackage{multirow}
\usepackage{makecell}
\usepackage{mathrsfs}
\usepackage[affil-it]{authblk}
\usepackage[colorlinks=true, allcolors=blue]{hyperref}
\usepackage{titles}
\usepackage{float} 
\usepackage{siunitx}

\usepackage{indentfirst}

\usepackage{abstract}

\newcommand\nnfootnote[1]{%
  \begin{NoHyper}
  \renewcommand\thefootnote{}\footnote{#1}%
  \addtocounter{footnote}{-1}%
  \end{NoHyper}
}

\title{Learning to Optimise Wind Farms with Graph Transformers}

\author[1]{\small Siyi Li$^*$}
\author[2]{\small Arnaud Robert}
\author[2,3,4]{\small A. Aldo Faisal}
\author[1]{\small Matthew D. Piggott}

\date{}

\affil[1]{Dept. of Earth Science \& Engineering,
Imperial College London, UK}
\affil[2]{Dept. of Computing,
Imperial College London, UK}
\affil[3]{Dept. of Bioengineering, Imperial College London, UK}
\affil[4]{Faculty VII, University of Bayreuth, Germany}

\begin{document}
\maketitle
\nnfootnote{$^*$Corresponding author's email address: siyi.li20@imperial.ac.uk}

%%=================================================%%
%%     Abstract                                         
%%=================================================%%
\begin{abstract}
This work proposes a novel data-driven model capable of providing accurate predictions for the power generation of all wind turbines in wind farms of arbitrary layout, yaw angle configurations and wind conditions. The proposed model functions by encoding a wind farm into a fully-connected graph and processing the graph representation through a graph transformer. The graph transformer surrogate is shown to generalise well and is able to uncover latent structural patterns within the graph representation of wind farms. It is demonstrated how the resulting surrogate model can be used to optimise yaw angle configurations using genetic algorithms, achieving similar levels of accuracy to industrially-standard wind farm simulation tools while only taking a fraction of the computational cost.

\end{abstract}

%%=================================================%%
%%     Introduction                                         
%%=================================================%%
\section{Introduction}  \label{introduction}

Wind energy is of immense importance in the global effort to address critical environmental, economic, and energy security challenges. Within wind farms, which consist of clusters of wind turbines strategically placed within confined areas, each individual wind turbine can significantly impact the performance of others due to the formation of a wake region downstream of its rotor. This wake effect is characterised by a reduction in wind speed due to the extraction of kinetic energy by the turbine rotor, as well as an increase in turbulence intensity generated through interactions between the airflow and the wind turbine structure, including its rotating blades \cite{les_paper}. When a downstream wind turbine operates within the full or partial wakes generated by the upstream turbines, it can experience a significant decrease in energy production. This occurs because airflow in the wake region has not yet fully returned to its undisturbed state, resulting in reduced wind speeds and increased turbulence \cite{NTNU_BT1, porte_agel_review}.
Furthermore, the heightened turbulence and shear within the wake also lead to an increase in dynamic loads on the turbine blades. As a consequence, the modelling of wind turbine wakes has garnered significant attention as one of the most critical elements in the optimal design and operational control of wind farms. For more in-depth insights, one can refer to \cite{PiggottReview2022} and the references therein.

One way to improve the performance of wind farms and mitigate the adverse effects of wake interactions is through yaw steering optimisation. Yaw steering optimisation introduces reasonable yaw angle adjustments to each individual wind turbine in order to help direct the wake from that turbine away from its downstream turbines. This is performed in a manner that seeks to enhance the power generation of the downstream turbines at the expense of a relatively minor reduction in power output from the upstream turbine in question. Furthermore, this needs to be performed for all turbines and needs to account for the hydrodynamic coupling between all turbines, i.e. optimisation cannot be performed sequentially for each turbine in turn, but needs to be formulated as a single `global' optimisation problem over the entire farm. In this manner, yaw steering optimisation functions as a strategic approach for optimising the overall performance of a wind farm by redistributing the wake impacts and maximising the collective power output. This contrasts with conventional control methods where wind turbines are oriented with the wind direction to maximise their own individual power generation rather than the whole farm. Numerous studies have demonstrated the effectiveness of yaw steering optimisation in improving wind farm power production. For instance, Adaramola and Krogstad \cite{ntnu_exp} conducted wind tunnel experiments with two lab-scale wind turbines, demonstrating a $12\%$ increase in power output through appropriate yaw adjustment of the upstream turbine. Ma et al. \cite{yaw_les} validated a cooperative yaw control strategy on five aligned wind turbines with high-fidelity large eddy simulation (LES), which showed an improvement in the power production of $17.5\%$. Howland et al. \cite{howland_pnas} reported a power production increase of up to $13\%$ through wake steering optimisation in an operational wind farm under specific wind conditions.

High-fidelity computational fluid dynamics (CFD) simulations, such as the Reynolds-Averaged Navier-Stokes (RANS) or LES approaches, are often used by researchers to accurately model the wake interactions between wind turbines under yawed conditions and estimate their impact on wind farm power generation. Nevertheless, the significant computational expenses and execution time associated with these methods often hinder their practical application within iterative design optimisation and as the underpinning for active control tools, especially for large-scale wind farms. On the other hand, analytical, semi-analytical, or so-called engineering wake models, including the Jensen model \cite{jensen_model}, the Larsen model \cite{larsen_model}, and the Bastankhah-Gaussian model \cite{gauss_model} are commonly integrated into industrial-standard software tools such as FLOw Redirection and Induction in Steady State (FLORIS) \cite{floris} and PyWake \cite{pywake} due to their rapid execution speed. However, despite being computationally efficient and able to provide quick estimation of wake behaviours, these models cannot fully capture the complex spatial and temporal evolution of the flow field and are highly dependent on empirical parameters and coefficients carefully calibrated against real or simulated data. 

Data-driven wake models, particularly those based on deep learning and trained on high-quality wind farm data, hold significant promise in balancing computational cost and accuracy. Many recent studies that integrate deep learning into wind farm power prediction rely on the establishment of surrogate models for individual wind turbine wakes. For instance, Ti et al. \cite{ann_rans} trained artificial neural networks (ANNs) on RANS data and employed wake superposition methods to effectively combine predicted individual turbine wakes and accurately calculate power. Zhang et al. \cite{gans_les} introduced an innovative wind turbine wake model capable of delivering 2D predictions by training a convolutional conditional generative adversarial neural network on LES data. Furthermore, Li et al. \cite{gnn_gad} utilised graph neural networks (GNNs) to learn from RANS simulations using their native multi-scale unstructured mesh based data, leading to the development of a data-driven wake model that can rapidly generate precise 3D flow field solutions. 

On the other hand, for the representation of entire farms, Park et al. \cite{pgnn_kaist} devised a data-driven model by representing wind conditions and relative positions of wind turbines as a graph and training a graph neural network with physics-induced attention (PGNN) that is able to predict individual turbine powers and the overall farm power accurately. Duthé et al. \cite{gnn_pywake} explored similar concepts by experimenting with various graph connectivities and assessing their impact on model performance. Additionally, Bentsen et al. \cite{gnn_attn} incorporated an attention mechanism that is entirely learned from training data, which adds more flexibility in weighing the importance of neighbouring vertices, leading to an improvement in model performance. Nevertheless, the representation of wind farms as graphs is not without limitations. One potential drawback is the reliance on pre-defined graph connectivities or adjacency matrices, which are crucial for model performance but can be hard to determine accurately.
Moreover, as the underlying physics varies significantly between models of different fidelities (e.g. analytical models vs CFD models vs real world data), GNNs trained with pre-established connectivities on a dataset of one may encounter difficulties in transfer learning to datasets of different fidelities. Additionally, the presence of discrete edges in the graph may introduce gradient discontinuities when the data-driven differentiable surrogate model is used to perform gradient-based optimisation of farm design parameters such as turbine layout and yaw angles. The work presented here aims to develop a more generic and generalisable framework for learning farm-scale data-driven models. Specifically, this work embraces a graph transformer approach, which offers the advantages of handling graph-structured data, as seen in GNNs, while benefiting from the impressive scalability and model capacity inherent to transformers \cite{transformer_paper}. This approach also mitigates the issues of over-smoothing \cite{over_smoothing}, which can complicate the construction of deep GNNs.

% Key contributions
%----------------------------------------------
\noindent \newline The primary novelties and contributions of this work can be summarised as follows:

\begin{enumerate}
    \item A novel farm-scale model based on graph transformers was developed using training data generated from PyWake. The model can very accurately predict wind farm power, averaging an accuracy of $99.8\%$ on layouts and wind conditions previously unseen during training.
    \item The developed data-driven model is able to precisely identify sparse wake interactions among turbines in the farm, i.e. to accurately characterise where there is aerodynamic influence between turbines and where there is none. This is achieved by the graph transformer while taking as input fully-connected graph representations of the wind farm containing information that only includes farm layouts, turbine yaw angles and wind conditions. This provides transparency into the inner workings of the model and aids in model interpretability.
    \item To improve the model's generalisability over diverse layouts and yaw angle configurations, a layout generator was employed to produce realistic wind farm layouts of different scales and sizes. Genetic algorithms (GAs) that search over various yaw angle configurations while aiming to optimise overall wind farm power were additionally utilised in the data generation process. Training data was randomly sampled from GA runs on wind farms of diverse layouts and wind conditions, enabling the model to discern between favorable and sub-optimal yaw angle configurations. The trained model is able to accurately identify optimal yaw angle configurations despite not encountering them during training.
    \item The performance and generalisability of the graph transformer model were demonstrated by utilising it as a surrogate model for PyWake in the context of yaw steering optimisation using GAs. The proposed model was shown to be able to generalise to arbitrary layouts and yaw angle configurations and can greatly expedite GA runs by grouping all offspring within a GA generation into a single batch.
\end{enumerate}

By providing an interpretable data-driven model that generalises to arbitrary wind farm layouts and yaw angle configurations, this work empowers industry practitioners with a rapid and reliable tool for optimising wind farm power. The significance of this research could be further amplified by subsequent investigations involving training or transfer learning on wind farm simulations featuring more intricate physics, such as that obtained from RANS or LES calculations, and/or by utilising real-world wind farm data.

The subsequent sections of this article are organised as follows: A brief summary of the background context for this work, encompassing engineering tools for wind farm simulation, GNNs and transformers are presented in Section \ref{background}. The proposed deep learning frameworks are introduced in Section \ref{methodology}. Detailed data generation processes and training experiments are presented in Section \ref{training}. Extensive performance tests and applications of the trained surrogate model are detailed in Section \ref{results}. Conclusions and potential future research directions are discussed in Section \ref{conclusions}.

%%=================================================%%
%%     Background                             
%%=================================================%%
%% broad literature review

\section{Background} \label{background}

\subsection*{Wind farm simulations with engineering wake models}
Engineering wake models are computationally inexpensive tools for computing the wake interactions within a wind farm across a wide range of steady-state conditions. This work makes use of the Bastankhah-Gaussian deficit model \cite{gauss_model} to calculate the velocity at each downstream wind turbine due to their upstream counterparts, and the Jiménez deflection model \cite{jimenez_model} to characterise the wake deflection of a wind turbine due to yaw steering. Both models are integrated into the open-source wind farm simulation software PyWake, which has been benchmarked against established high-fidelity simulations and real-world wind farm data \cite{pywake, pywake_validation}. The velocity deficit with an assumed Gaussian shape can be written as

\begin{align}
        &\Delta U = \bigg( 1 - \sqrt{1 - \frac{C_\text{T}}{8 \big( {\sigma(x)/d_0}^2 \big) }} \bigg) \text{exp} \bigg(-\frac{x^2}{2{\sigma(x)}^2} \bigg) U_{\infty},
\end{align}

\noindent where $x$ is the downstream distance from the turbine rotor, $\sigma(x)$ is the standard deviation of the Gaussian velocity deficit profile at $x$, $C_{\text{T}}$ is the thrust coefficient of the turbine, $U_{\infty}$ is the incoming wind velocity and $d_0$ is the wind turbine's diameter. The standard deviation, $\sigma(x)$, is commonly defined to be
\begin{align}
    &\sigma(x) = k x + 0.2 \sqrt{\beta} d_0,
\end{align}
where $k = \frac{\partial \sigma(x)}{\partial x}$ is the growth rate of the wake and $\beta$ is a function of $C_{\text{T}}$.

After being amended with lateral deflection of the wake due to the Jiménez model, a superposition-based approach is then used to compute the effective wind speed $U_{\text{w}}$ at each turbine, given the incoming wind speed and wake deficits from upstream turbines. The power of a wind turbine can then be found as
\begin{align}
    &P = \int_{A_0} \frac{1}{2} \rho U^3_w \, dA,
\end{align}
where $A_0$ is the area swept by the wind turbine blades and $\rho$ is the density of air.

\subsection*{Graph neural networks and transformers}
A wide variety of physical, biological and social systems can be formulated as graphs at a certain level of abstraction. The utilisation of graph representations in these systems facilitates the applications of various mathematical and computational tools, including graph neural networks (GNNs), allowing researchers to uncover hidden relationships, make accurate predictions and solve complex problems. While traditional machine learning techniques that are effective across many scenarios often prove inadequate when applied to graph-structured data, GNNs have had great success in various domains where relationships and dependencies are embedded within network structures. Recent years have witnessed the application of GNNs in predicting material properties of crystals \cite{gnn_molecule}, identifying influential users within online social networks \cite{gnn_social}, recommend meaningful mobile applications to potential users \cite{gnn_mobile} and simulating complex mesh-based numerical models of high-dimensional physical systems \cite{meshgraphnet}. The transformer model \cite{transformer_paper} has been widely acknowledged as the most powerful neural network for modelling sequential data, and has seen great success in natural language processing (NLP), speech analysis \cite{trans_speech} and computer vision \cite{trans_swin}. In recent years, researchers have found ways to alleviate limitations pertaining to GNNs such as over-smoothing \cite{over_smoothing}, over-squashing \cite{over_squashing} by incorporating some of the principles underpinning transformers, but adapting them to modelling the unique properties of graphs \cite{graphormer, graph_gps}. More details about GNNs and graph transformers are introduced in Section \ref{methodology}.

%%=================================================%%
%%     Methodology                             
%%=================================================%%
\section{Methodology} \label{methodology}

\subsection*{Graph representation of wind farms}
The representation of a wind farm, along with wind direction information, can be encoded as a directed graph denoted by $\mathcal{G} = (V, E, u)$.  Here, $\mathcal{G}$ is a tuple consisting of vertex-level features denoted as $V \in \mathbb{R}^{N_v \times d_v}$, edge-level features represented as $E \in \mathbb{R}^{N_e \times d_e}$, and graph-level features expressed as $u \in \mathbb{R}^{d_u}$. In this context, $N_v$ and $N_e$ refer to the number of vertices and edges, while $d_v$, $d_e$, and $d_u$ indicate the dimensions of vertex, edge, and graph-level features, respectively. Each vertex, labelled as ${v_i}$ for ${i=1},\ldots,{N_v}$ within the graph, corresponds to one of the $N_v$ wind turbines present in the farm. The vertex-level features encompass information such as the turbine's position and yaw angle. The edges in the graph signify wake effects or the influence on power production of downstream turbines due to those upstream. A directed edge, denoted as $e_{ij} \in E$, is established if the wake generated by $v_i$ may influence the power output of $v_j$. The specific process for determining these edges in the graph representation of wind farms is detailed in Section \ref{methodology}. Edge-level features could encompass factors such as relative positions and distances between turbines. Additionally, graph-level features incorporate physical parameters shared among all vertices in the graph, including variables such as wind speed and turbulence intensity (TI).

This approach of representing wind farms as graphs drew inspiration from the research conducted by Park et al. \cite{pgnn_kaist} and assumes that the interactions captured within the graph provide an adequate basis for forecasting the wind farm's yield. More details of the wind farm graph representation are given in Section \ref{methodology}. It should be noted that the graph representation of wind farms in this context exhibits pronounced heterophily, where edges encode dissimilarity in power attributes rather than similarity. Consequently, many industrial-standard graph neural network models, such as the Graph Convolutional Network (GCN) \cite{gcn_paper}, which are primarily designed for modelling homophilous graphs, may not be the optimal choices for addressing this specific learning task. On the other hand, attentional networks that aggregate neighbouring vertex features using implicit weights through attention or graph neural networks with a message-passing approach capable of computing versatile vectors or messages for transmission along edges tend to be more appropriate.

\subsection*{Message-passing graph neural networks}
The message-passing graph neural network \cite{battaglia_gn} represents one of the most versatile variations of graph neural networks, allowing for the transmission of arbitrary message vectors between vertices connected by edges. In this framework, edges in a graph provide a blueprint for transmitting information. The content of the information passed is determined by a message function that relies on the features of both the sender and receiver vertices. 

\begin{figure}[H]
    \centering
    \includegraphics[width=\textwidth]{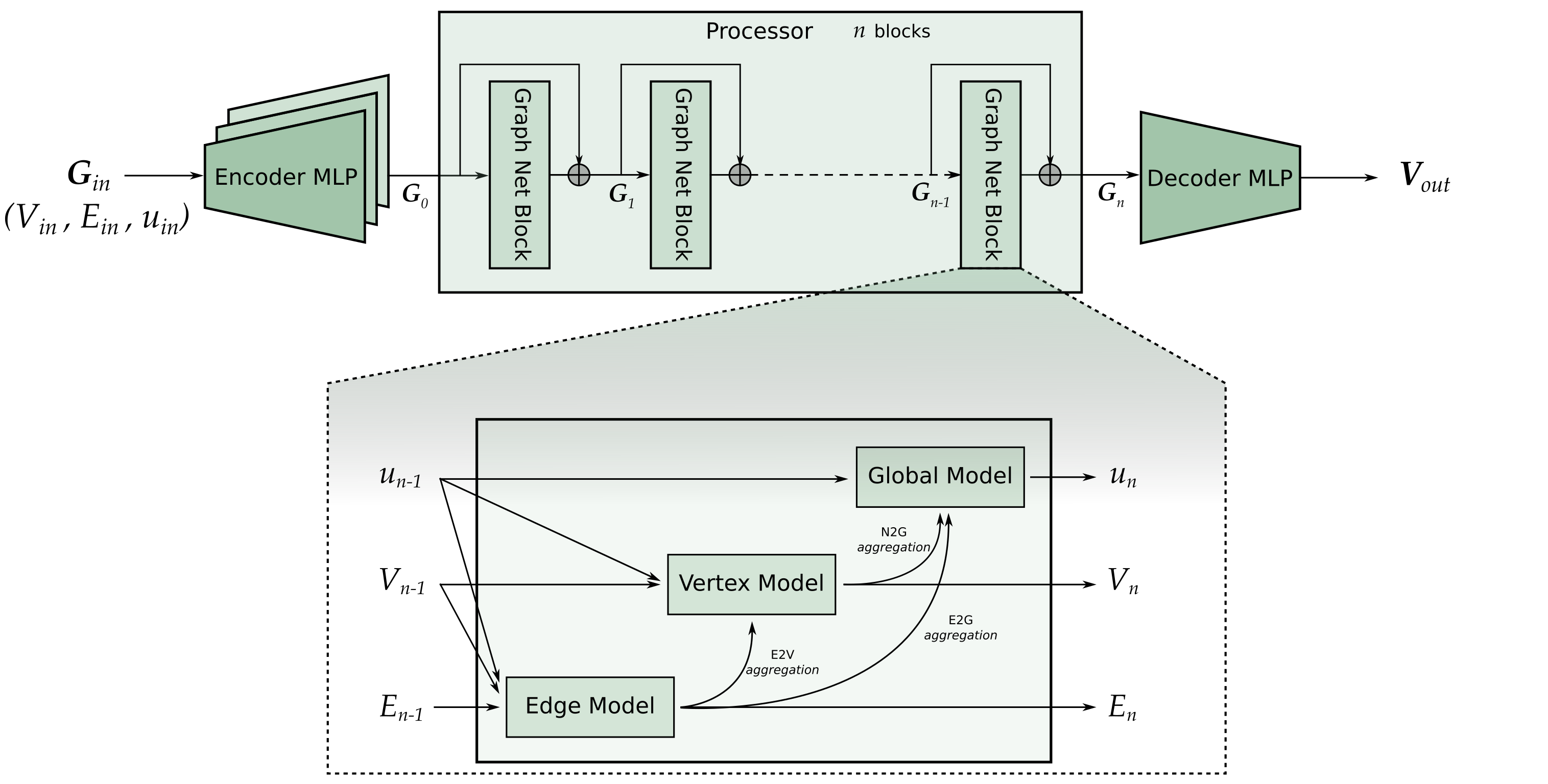}
    \caption{Schematic of the message-passing graph network architecture. The edge, vertex and global models in a graph net block are multi-layer perceptrons (MLPs).}
    \label{fig:gn_illus}
\end{figure}

The graph neural network architecture used in this work closely aligns with the designs detailed in the papers by Battaglia et al. \cite{battaglia_gn} and the work by Park et al. \cite{pgnn_kaist}. In the context of a wind farm consisting of $N$ turbines, the graph neural network model receives input data $\mathcal{G} = (V, E, u)$, where $V \in \mathbb{R}^{N_v \times d_v}$. In the work presented here, $N_v$ is considered to range from 2 to 100, representing the number of wind turbines, and $d_v = 3$ includes turbine-level features such as span-wise and stream-wise coordinates and turbine yaw angles. The edge features $E \in \mathbb{R}^{N_e \times 2}$ encompass the relative span-wise and stream-wise distances between turbines, while the global features $u \in \mathbb{R}^{2}$ comprise the farm's upstream wind speed and turbulence intensity. The GNN architecture comprises three MLP encoders, each tasked with transforming vertex, edge, and global features into more expressive latent spaces. These encoded features are subsequently processed through a series of graph net blocks. Within each graph net block, an edge model combines edge features, source and target vertex features, and global features to generate new edge embeddings. A vertex model aggregates incoming edge features with existing vertex and global features to produce new vertex embeddings. Finally, a global model constructs new graph-level embeddings by pooling all vertex and edge-level features, concatenated with the global features. A decoder MLP then utilises the updated latent representation computed by the processor to extract meaningful information. In this case, the transformed vertex features are decoded to generate vertex-level predictions for individual turbine powers. An illustration of the message-passing graph neural network is shown in Figure \ref{fig:gn_illus}. 

\subsection*{Graph transformers}

\begin{figure}[!b]
    \centering
    \includegraphics[width=\textwidth]{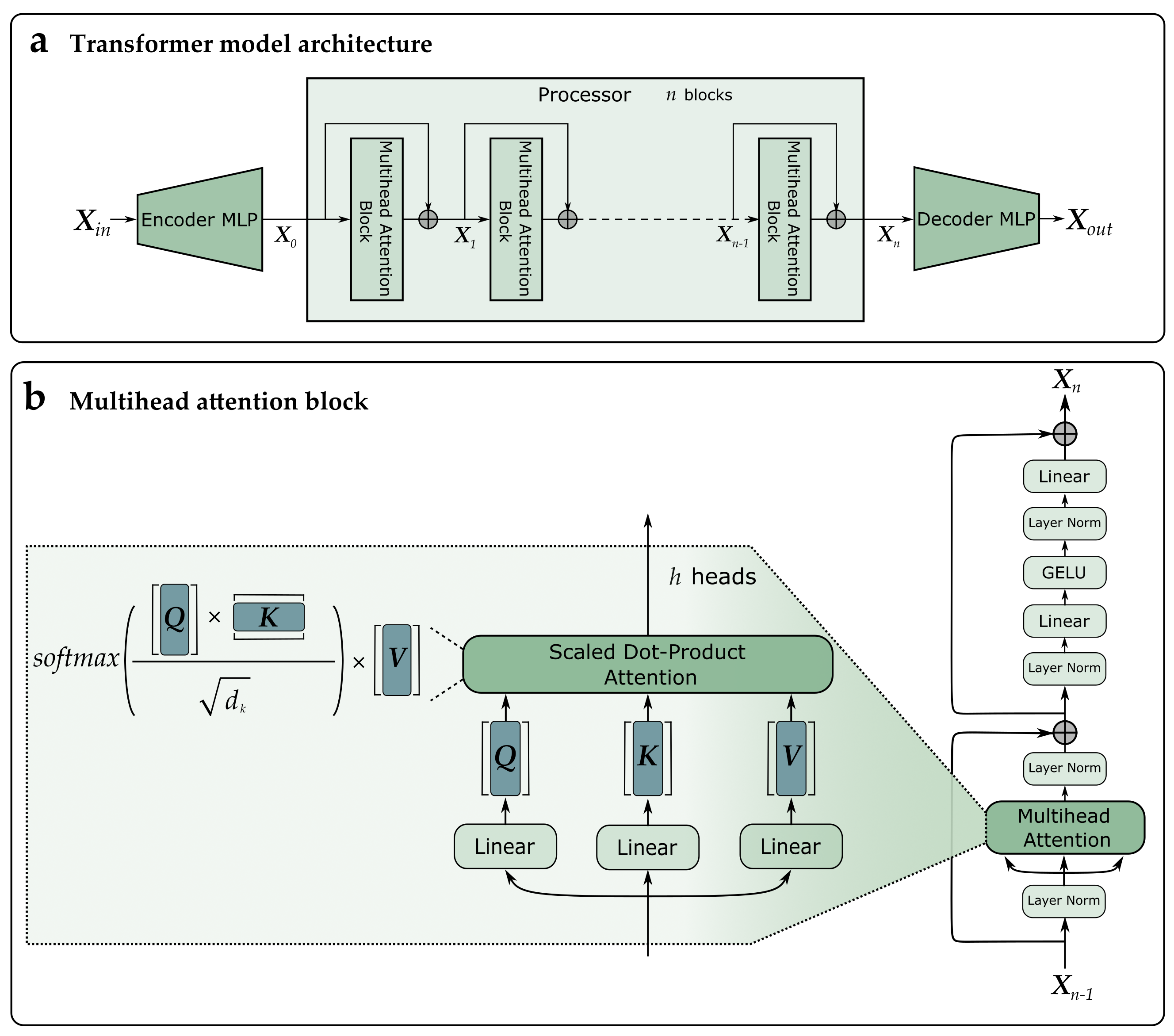}
    \caption{Schematic of the proposed transformer model architecture. $\mathbf{a}$: The proposed transformer architecture and its building blocks. $\mathbf{b}$: Details of a multi-head attention block, where GELU refers to the Gaussian Error Linear Unit.}
    \label{fig:trans_architecture}
\end{figure}

The architectural design of the proposed graph transformer adheres to the encoder-processor-decoder structure, where the graph net blocks are substituted with multi-head attention blocks. These multi-head attention blocks draw inspiration from those in the encoder section of the original transformer paper \cite{transformer_paper} and incorporate elements from the transformer architecture with additional layer normalisation, as described in \cite{normformer} and \cite{particle_trans}. In the context of a wind farm comprising $N$ turbines, the transformer model takes as input only vertex-level features $X \in \mathbb{R}^{N \times 5}$, where $N \in [2, 100]$ denotes the number of wind turbines. Note that the 100 quoted here represents an upper limit on the number of turbines with a farm, and which could be readily increased. The vertex-level features have global features appended to them, which include the span-wise and stream-wise coordinates of the wind turbines, turbine yaw angles, and the upstream wind speed and turbulence intensity. In particular, no explicit positional encodings were applied, aligning with the principle that wind farm graph representations should exhibit permutation invariance while acknowledging that spatial coordinates were directly incorporated into the input features.

The multi-head attention block within the proposed architecture is a key component designed to efficiently learn and depict the interactions among wind turbines in the farm. This is crucial for achieving accurate power predictions and gaining valuable insights into the patterns of wind farm power generation. Each multi-head attention block comprises two fundamental components: a multi-head self-attention module and a point-wise feed-forward network. The multi-head self-attention module empowers the model to weigh and process vertex-level features differently, capturing nuanced relationships and dependencies among the wind turbines in the farm. It enhances the model's ability to understand the intricate interactions between turbines, such as the impact of their relative positions, orientations, and physical conditions on power generation. The point-wise feed-forward network further refines the learned representations by applying non-linear transformations to each feature independently. Incorporating additional layer-normalisation layers into the architecture further enhances model stability, contributes to faster convergence, and bolsters overall robustness. Figure \ref{fig:trans_architecture} offers a visual depiction of the model architecture, with a detailed focus on the multi-head attention block. Within each multi-head attention block, the input $X \in \mathbb{R}^{N \times d}$ is projected to query ($Q$), key ($K$) and value ($V$) representations by three matrices $W_Q \in \mathbb{R}^{d \times d_k}$, $W_K \in \mathbb{R}^{d \times d_k}$ and $W_V \in \mathbb{R}^{d \times d_v}$ respectively, where $d$ is the hidden dimension with $d_k = d_v = d$. The self-attention is then computed as

\begin{align}
        &Q = XW_Q,\;K=XW_K,\;V=XW_{V},\\
        &\text{Self-attention}(X) = \text{softmax}\bigg(\frac{QK^{\text{T}}}{\sqrt{d_k}}\bigg) V.
\end{align}

%%=================================================%%
%%     Model training                         
%%=================================================%%
\section{Model training} \label{training}

\subsection*{Data generation}
\begin{figure}[H]
    \centering
    \includegraphics[width=\textwidth]{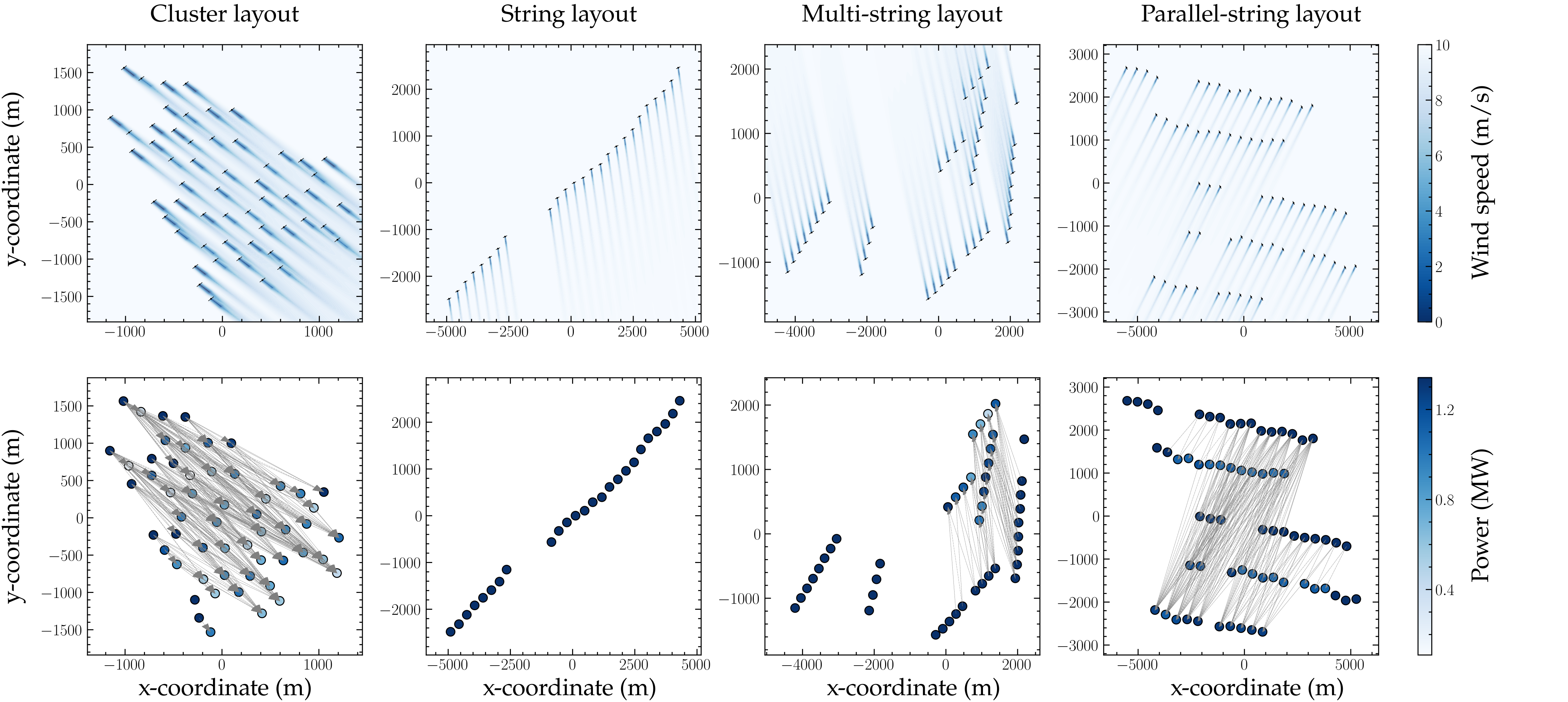}
    \caption{Visualisation of various layout styles produced by PLayGen and their corresponding graph representations. Wake maps were generated using the Bastankhah-Gaussian analytical model within PyWake. For the purposes of these visualisation, simulations were conducted with a wind speed of $10$ $m/s$, $5\%$ turbulence intensity, and all turbines facing the wind direction which in each case was chosen at random. It is important to note that flow field information was neither produced during the data generation process nor included as a component of the features within the dataset. The directed arrows represent edges used in GNN training. Wind turbines were coloured based on their individual power generation.}
    \label{fig:playgen_layouts}
\end{figure}

Wind farm layouts were randomly generated using an adapted version of the Plant Layout Generator (PLayGen), originally developed by NREL \cite{wpgnn_nrel}. Subsequent wind farm simulations were carried out using the PyWake \cite{pywake} software. The simulation model incorporated the Bastankhah-Gaussian analytical model to replicate wake effects \cite{gauss_model}. Individual turbine wakes were combined using the sum-of-squares superposition method, and the Jiménez wake deflection model was employed to simulate yaw angle based wake steering effects \cite{jimenez_model}. Graph representation of wind farms was constructed incorporating the farm layouts, simulation parameters and individual turbine powers obtained through PyWake simulations. For GNN training, directed edges were established from an upstream wind turbine to a downstream one if the angle between the two fell within the range of [$-15^{\circ}, 15^{\circ}$], the graph transformer instead opted for a fully-connected approach. Figure \ref{fig:playgen_layouts} provides visual depictions of various wind farm layouts generated by PLayGen, alongside their corresponding graph representations. Notably, the wind turbine parameters employed in all simulation scenarios were calibrated to align with the specifications of the Vestas V80 wind turbine.

Two datasets were generated for training and experimenting with different machine learning-based surrogate models. The ``standard'' dataset encompasses diverse wind farm layouts and physical parameters with randomly initialised wind turbine yaw angles, serving as a validation set for experimenting with various surrogate models. In contrast, the ``enhanced'' dataset is significantly larger and incorporates a broader range of yaw angle configurations. The standard dataset was utilised for initial model validation and experimentation, while the enhanced dataset is specifically designed to leverage the transformer models' capabilities to handle enormous datasets and train the most advanced surrogate model for wind farm power prediction and optimisation.

In the standard dataset, 200k simulation scenarios were generated, each scenario subjected to a random wind farm layout from one of the four layout styles shown in Figure \ref{fig:playgen_layouts}, with a random number of wind turbines between 2 and 100, wind speed between $8$ and $15$ $m/s$, wind direction between $0^{\circ}$ and $359^{\circ}$, turbulence intensity (TI) between $5\%$ and $15\%$. All wind turbine yaw angles were randomly initialised between $-30^{\circ}$ and $30^{\circ}$. In the enhanced dataset, on the other hand, to improve the model's ability to discern subtle variations in yaw angle configurations, the training data was randomly sampled from different genetic algorithm runs. Each GA run was performed using the PyWake model such as to optimise for total power generation by varying individual wind turbine yaw angles between $-30^{\circ}$ and $30^{\circ}$. All instances of the genetic algorithm were executed for 50 generations with a population size of 100, a cross-over probability of $70\%$ and a mutation rate of $50\%$ so that a more diverse range of yaw angle configurations can be introduced. A total of 200,000 scenarios (each comprising a unique wind farm layout and set of physical wind parameters) were randomly generated and optimised using genetic algorithms to produce the dataset. The final dataset included 100 randomly sampled yaw angle configurations from each scenario and was split into $80\%$ for training, $10\%$ for validation and another $10\%$ for testing. 

The creation of both datasets was parallelised and executed on Imperial College London’s CX1 High-Performance Computing (HPC) cluster. The generation of the standard dataset required approximately eight CPU hours, while the enhanced dataset necessitated over 5,000 CPU hours.

\subsection*{Supervised training experiments}
All models were trained to minimise the mean squared error (MSE) of predicted individual turbine powers with the AdamW optimiser \cite{adamw}. Training experiments and model comparisons were conducted primarily on the standard dataset, with a final larger model being trained on the enhanced dataset. 

\begin{figure}[H]
    \centering
    \includegraphics[width=0.5\textwidth]{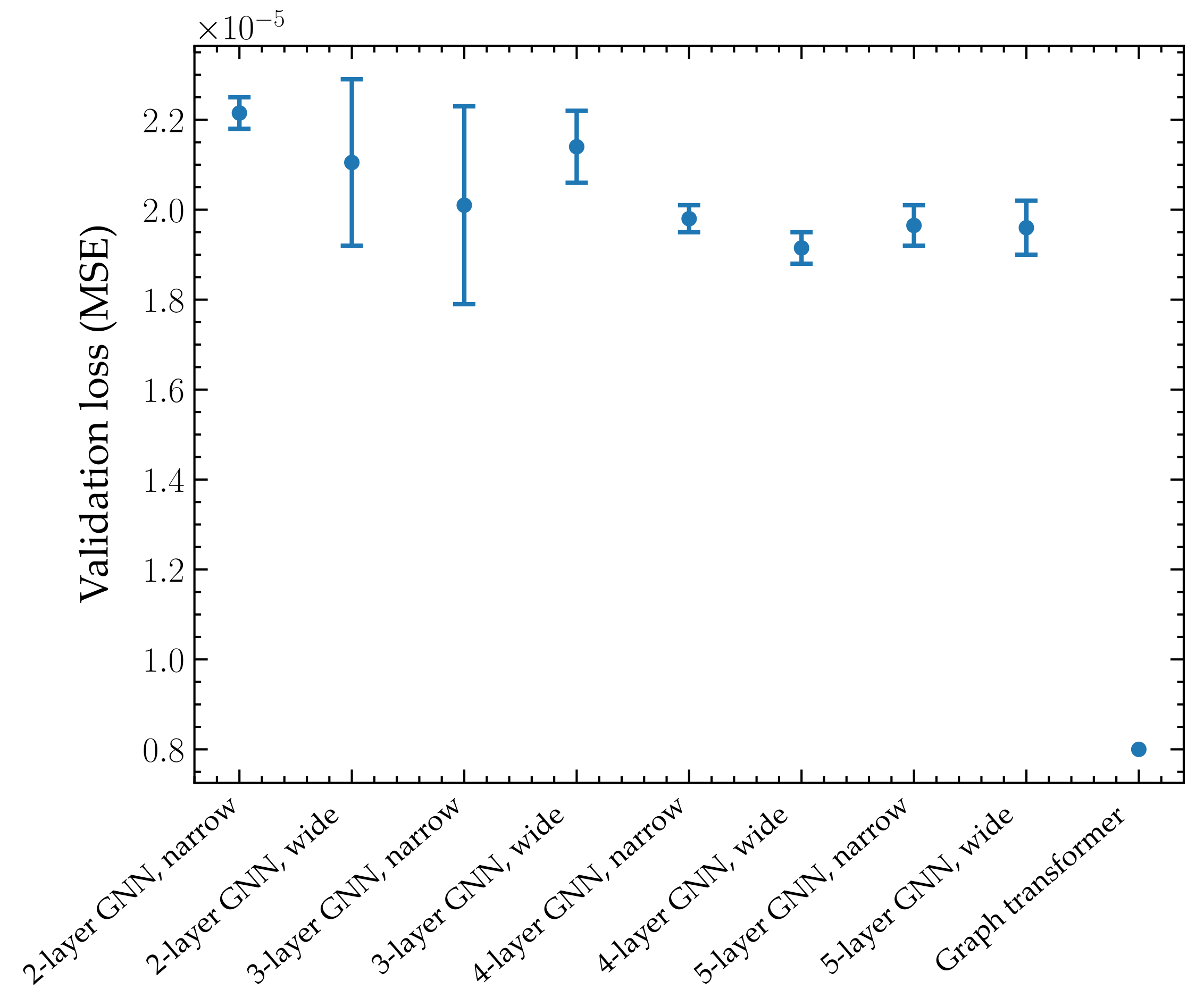}
    \caption{Comparison of lowest validation loss of GNN models of various sizes and the graph transformer model. The lowest validation loss of each GNN model was computed from training runs with three different random seeds. GNN models with 64 hidden units per MLP layer were denoted as ``narrow'', while those with 128 hidden units per MLP layer were referred to as ``wide''. The graph transformer model was trained only once due to the excessive training time required.}
    \label{fig:model_comparison}
\end{figure}

Figure \ref{fig:model_comparison} illustrates the results of training various GNN models with different numbers of graph net blocks and hidden units per Multi-Layer Perceptron (MLP) layer within the encoder, decoder, and graph net blocks. These are compared to a graph transformer consisting of five multi-head attention blocks, four attention heads, and a hidden size of 128. The graph transformer model featured a total of 16.3 million trainable parameters and reached its lowest validation loss at $8.6 \times 10^{-6}$. In contrast, the best-performing GNN model achieved a validation loss of $1.79 \times 10^{-5}$ at 124 thousand parameters. The results indicate that increasing model capacity by adding more layers and hidden units per layer led to only marginal improvements in GNN performance. On the other hand, the graph transformer model demonstrated superior scalability with respect to the number of trainable parameters in the model, and a sufficiently large graph transformer was able to outperform the GNNs in terms of validation loss.

The final choice for the surrogate model involved opting for a transformer model with increased width and depth. This model comprised eight multihead attention blocks, each with four attention heads of 128 hidden dimensions. The encoder and decoder MLPs both possessed 512 hidden dimensions, resulting in a model with a total of 25.8 million trainable parameters. The model was trained for one million steps on the enhanced dataset with a batch size of 256. A cosine annealing learning rate scheduler \cite{sgdr} was incorporated into the training process, with a maximum learning rate of $4\times10^{-4}$, which was reached after a linear warm-up period of 50,000 steps. To optimise GPU memory usage and accelerate training, 16-bit automatic mixed precision (AMP) was employed. The entire training process spanned 75 hours and was conducted on 5 NVIDIA A100 GPUs on the Department of Earth Science and Engineering's ``Hivemind'' DGX GPU cluster.

%%=================================================%%
%%     Results and discussion                        
%%=================================================%%
\section{Results and Discussion} \label{results}
\subsection*{Assessment of surrogate model efficacy}
\begin{figure}[H]
    \centering
    \includegraphics[width=\textwidth]{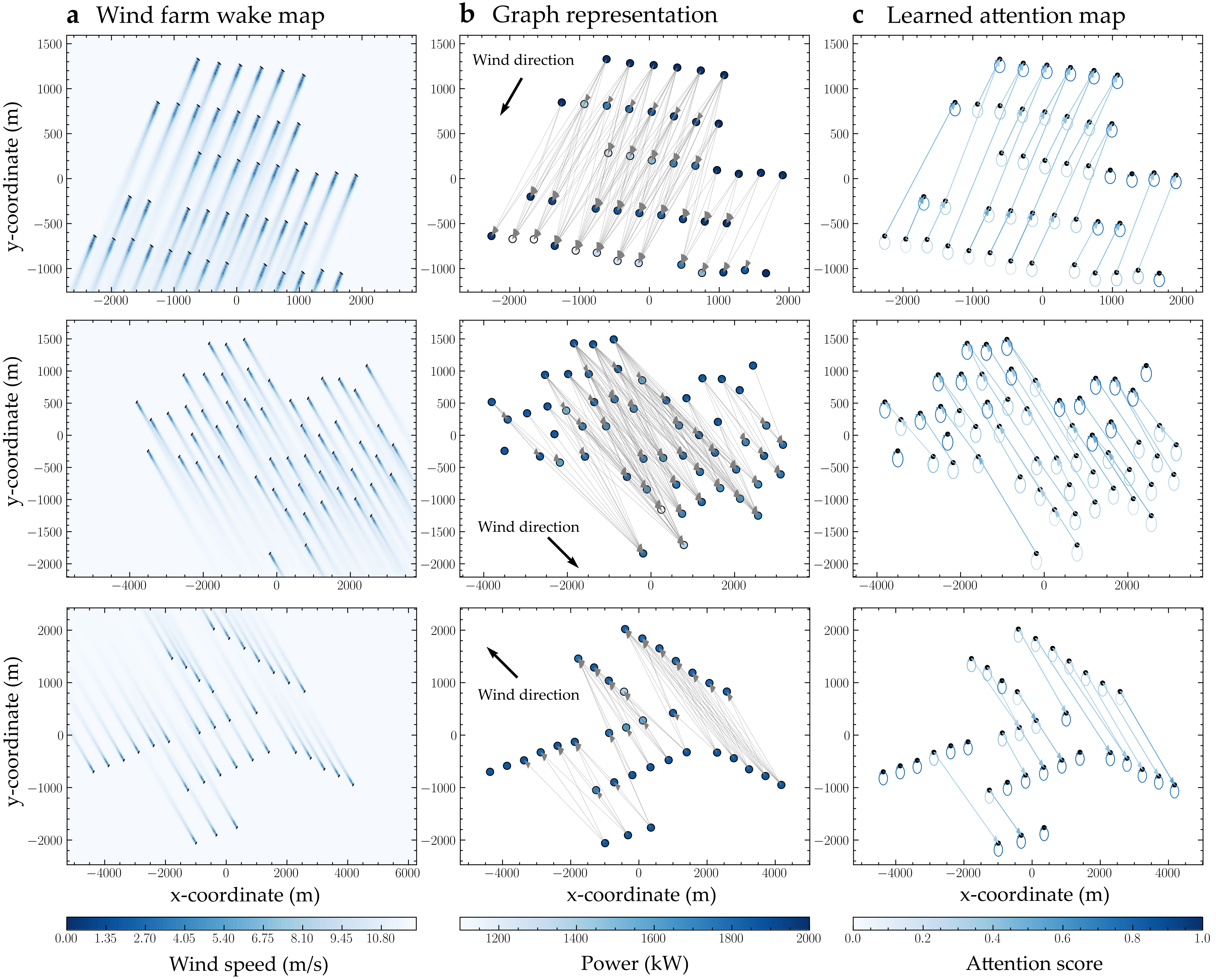}
    \caption{Illustration of different approaches for representing wind farms. The three rows correspond to three randomly generated wind farms previously unseen by the model, subject to similar physical conditions of $12$ $m/s$ wind speed and $5\%$ TI at incoming wind directions of $45^\circ$, $315^\circ$ and $135^\circ$. The three wind farms comprise 45, 55, and 35 wind turbines, respectively. In particular, all turbines in all configurations are orientated to face the direction of the wind. $\mathbf{a}$: Wake map produced by PyWake using the Bastankhah-Gaussian analytical model. $\mathbf{b}$: Graph representation of the wind farms, where solid dots represent wind turbines (vertices) and grey lines with arrows represent interactions (directed edges). $\mathbf{c}$: Graph representation of the attention score output from the trained transformer model. Only attention scores larger than 0.1 are visualised. Circles with arrows at the end denote self-attending tokens.}
    \label{fig:farm_rep}
\end{figure}

As mentioned in Section \ref{methodology}, a wind farm can be naturally defined as a graph, where wind turbines are represented as vertices of the graph, and the wake effects among turbines can be modelled as (directed) edges between pairs of vertices. Figure \ref{fig:farm_rep}b illustrates graph representations of randomly generated wind farms under different incoming wind directions. The graph representations were created by treating the wind turbines as graph vertices and establishing edges based on wind direction, where a directed edge is established from an upstream turbine to its downstream counterparts. This edge is formed when the angle between the pairs of turbines and the wind direction falls within a range of $15^{\circ}$, considering potential yaw steering effects. It is important to acknowledge that alternative methods for determining edges, such as nearest-neighbour connections or fully connected edges, are equally valid. 

In situations where graph learning is applied to problems lacking predefined relationships among vertices, the approach chosen for defining edges, be it based on factors such as wind direction or downstream distance, markedly impacts the effectiveness of the graph learning tasks. As sub-optimal choices of edges could hinder the predictive power and generalisability of the model, one potential alternative is to consider fully connected graphs, which is similar to the graph transformer approach adopted here, where each vertex is free to interact with every other vertex and the significance of this pair-wise interaction is evaluated through an attention score.

Attention scores computed within a transformer model can enhance model interpretability by offering valuable insights into how the model focuses on different parts of the input data to make accurate predictions. Graph representations of the attention map output from the surrogate model on previously unseen wind farms are shown in Figure \ref{fig:farm_rep}c. The attention map was extracted from the last multi-head attention block in the transformer model and averaged over all attention heads.

It can be seen that the transformer model is able to demonstrate a notable proficiency in accurately capturing the dependencies of power and wake effects among wind turbines, even when flow field data was not directly provided as model input. Moreover, there is a consistent correspondence between the attention scores learned by the graph transformer model and the wake patterns deducible from the wind farm's wake map produced (for visualisation purposes) by PyWake.

\begin{figure}[H]
    \centering
    \includegraphics[width=\textwidth]{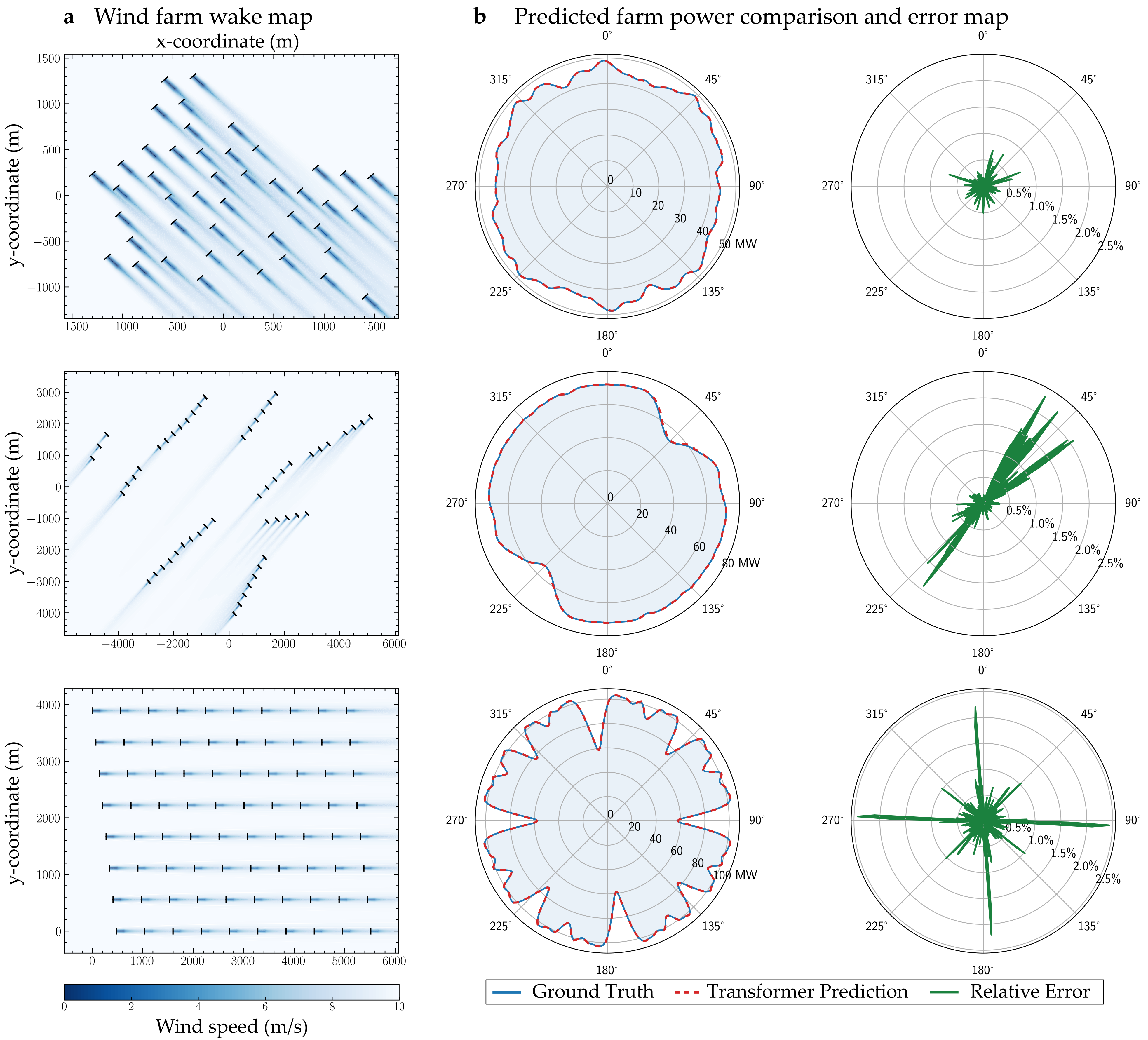}
    \caption{Model testing was conducted across various wind directions using previously unseen farm layouts. The results are presented in three rows, demonstrating outcomes for a random cluster-type layout, a random multi-string-type layout, and the Horns Rev 1 layout, respectively. $\mathbf{a}$: Wake map generated by PyWake using the Bastankhah-Gaussian analytical model. The assumed conditions for the three wind farms included wind speeds of $10$ $m/s$, TIs of $5\%$, and dominant wind directions of $315^{\circ}$, $45^{\circ}$, and $270^{\circ}$, respectively. All wind turbines were oriented to face the incoming wind direction. $\mathbf{b}$: Comparison of total farm power computed by PyWake and the transformer surrogate model, as well as relative error at different wind directions.}
    \label{fig:wind_dir}
\end{figure}

The assessment of the transformer surrogate model's predictive power and its ability to generalise was extended by employing it to predict the power output of wind farms with layouts not encountered during the training phase. This evaluation spans all wind directions ranging from $0^{\circ}$ to $359^{\circ}$ at intervals of $1^{\circ}$. Three exemplar outcomes are showcased in Figure \ref{fig:wind_dir}, where the farm layouts are presented in Figure \ref{fig:wind_dir}a. Figure \ref{fig:wind_dir}b illustrates the predicted farm power from PyWake, regarded as the ground truth for the purposes of this work, the predictions from the transformer surrogate, and the discrepancy between the two. The first two rows of Figure \ref{fig:wind_dir} show the evaluation of the model on two randomly generated layouts featuring 45 and 55 wind turbines, respectively. The third row shows an evaluation of the model on a real-world wind farm, Horns Rev 1, a Danish offshore installation of 80 wind turbines situated in the North Sea. 
Across all wind directions, the transformer surrogate demonstrates average mean relative accuracies of $99.85\%$, $99.71\%$, and $99.69\%$, along with minimum accuracies of $99.25\%$, $97.65\%$, and $97.56\%$, for the cluster, multi-string, and Horns Rev 1 layouts, respectively. In general, the slightly reduced accuracies tend to correspond to wind directions associated with more rapid changes in total wind farm power. This is explainable in part as we plot relative error values, and the related fact that higher model errors corresponding to those wind directions where wake effects between turbines are strongest. The strong agreement observed between the predictions made by the transformer surrogate and PyWake serves as additional evidence of the model's generalisability. Remarkably, even without explicit training on identical layouts across different wind directions, the transformer model is able to demonstrate high accuracy when applied to novel layouts under different wind directions.

\subsection*{Yaw optimisation with transformer model and genetic algorithm}

The robustness and generalisation capabilities of the transformer surrogate model in predicting farm power were thoroughly assessed through its application in optimising yaw angle configurations using genetic algorithms. The model underwent testing under more demanding conditions, characterised by GA runs with an increased number of individuals per generation and an extended evolutionary timeline. These challenging test conditions, featuring a cross-over rate of $70\%$, a mutation rate of $10\%$, and 200 individuals per generation over 500 total generations, marked a substantial increase from the 100 individuals for 50 generations used in the data generation process. This deliberate escalation was aimed to scrutinise the model's adaptability to heightened problem complexity. The enlarged search space from a larger pool of individuals and extended number of generations, confronts the model with a more extensive exploration of potential yaw angle configurations, and tests the model's capacity to discern relevant relationships among wind turbines within wind farms and to accurately predict powers generated from wind farms with arbitrary layouts and yaw angle configurations. Moreover, the extended evolutionary timeline introduces additional long-term dependencies in the GA runs, which necessitates the surrogate model to adapt to evolving dynamics during the optimisation process and remain stable and accurate over an extended duration. Importantly, the increased number of generations in GA runs also tests the surrogate model's ability to generalise to highly optimised yaw angle configurations not encountered during training.

\begin{figure}[h]
    \centering
    \includegraphics[width=\textwidth]{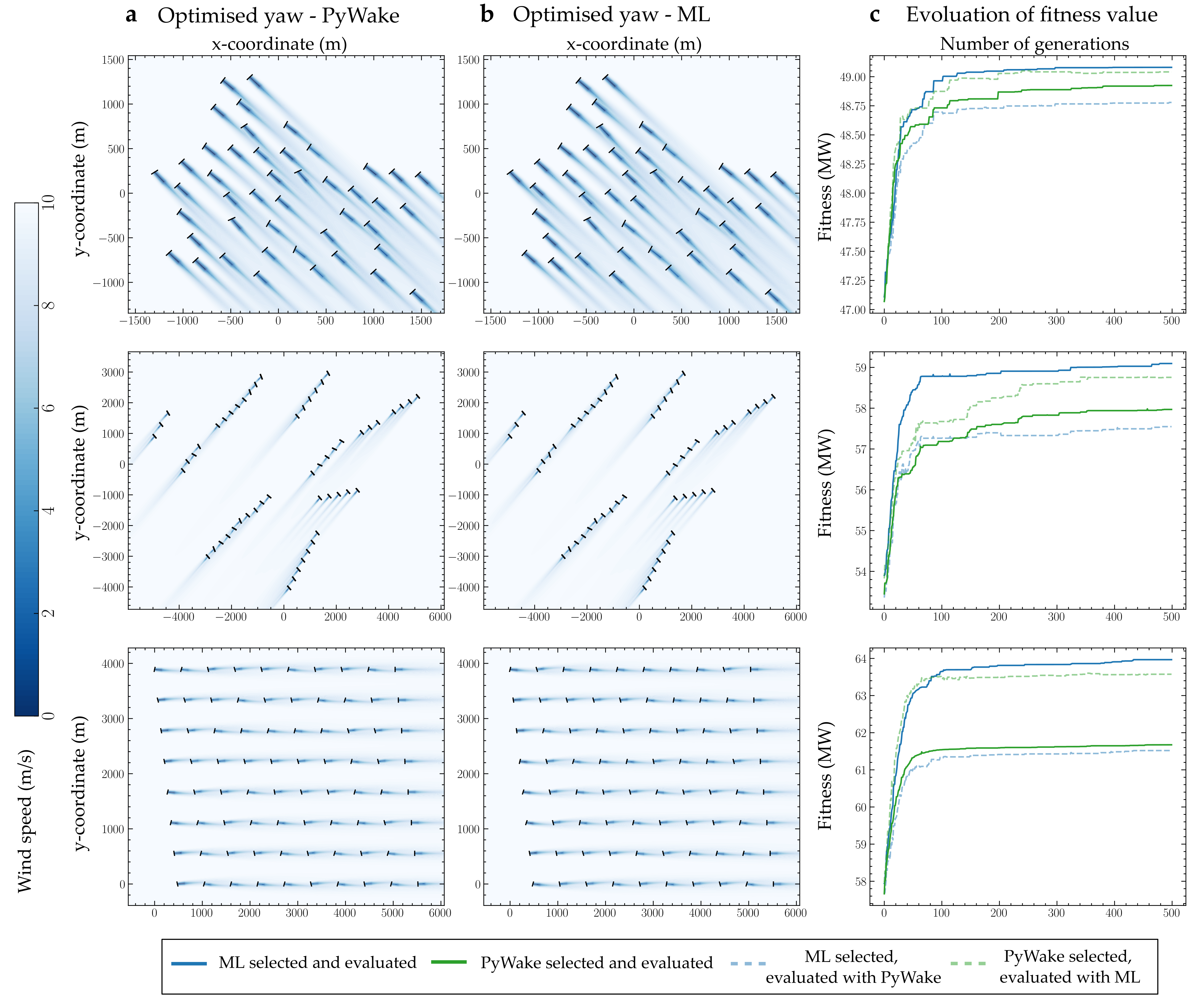}
    \caption{Results of genetic algorithm runs with PyWake and the transformer surrogate model and their comparisons. $\mathbf{a}$: Wake maps featuring optimal yaw angle configurations identified through GA runs with PyWake. $\mathbf{b}$: Wake maps (obtained using PyWake) showcasing optimal yaw angle configurations from GA runs with the transformer surrogate model.  $\mathbf{c}$: Evolution of the top-performing individual in in each genetic algorithm generation. Dashed lines denote the fitness of the leading individual selected by one model but evaluated using the other.}
    \label{fig:ga_runs}
\end{figure}

Three instances of genetic algorithm optimised yaw configurations, on wind farm layouts previously unseen by the surrogate model during training and were previously depicted in Figure \ref{fig:wind_dir}a, are compared and presented in Figure \ref{fig:ga_runs} for both PyWake and the transformer surrogate model.

The optimised yaw angle configurations in Figure \ref{fig:ga_runs}a and Figure \ref{fig:ga_runs}b highlight the transformer surrogate model's ability to effectively identify wind turbines for yawing, as the resulting wake maps from the top-performing individuals identified through PyWake and the transformer surrogate exhibit significant similarity. Despite the surrogate model's tendency to overestimate farm power in highly optimised individuals, the transformer predictions consistently maintain accuracy across successive generations of genetic algorithms, as illustrated in Figure \ref{fig:ga_runs}c. When comparing the top-performing individuals selected and evaluated by the transformer surrogate against those selected and evaluated by PyWake, the relative errors were $99.7\%$, $98.1\%$, and $96.3\%$ for the cluster, multi-string, and Horns Rev layout, respectively. Additionally, the surrogate model excels in discerning and selecting optimal yaw angle configurations, as evidenced by the high relative accuracies of top individual fitness. Specifically, when evaluating the top individual identified by the transformer model but assessed with PyWake, compared to the top individual selected and evaluated directly with PyWake, the relative accuracies for the three cases were $99.7\%$, $99.3\%$, and $99.8\%$, respectively. 

Given the surrogate model's ability to process a batch of various yaw angle configurations, running genetic algorithms with the surrogate model can be significantly expedited by grouping all individuals in a generation into a single batch and using it as input. Genetic algorithm runs conducted with PyWake on the three exemplary wind farms in Figure \ref{fig:ga_runs} featuring 45, 55, and 80 wind turbines, where individuals in a generation were evaluated separately, required 25.5, 30.4, and 44.3 minutes, respectively. In contrast, genetic algorithm runs with the transformer surrogate with batched individuals were completed in approximately one minute for all three cases. Notably, the evaluation time of the surrogate model remains constant regardless of the number of turbines, as predictions are made using dense batches of size 100. Hence, given the insights from the prior examinations of the surrogate model's performance in Figure \ref{fig:ga_runs}c, and considering its very rapid execution time, especially when combined with GA, one viable option would be to exclusively leverage the surrogate model for identifying the optimal yaw angle configurations. The final assessment of this chosen configuration can then be performed using PyWake. This streamlined approach then has the potential to substantially decrease the overall simulation time.

%%=================================================%%
%%     Conclusions                        
%%=================================================%%
\section{Conclusions} \label{conclusions}
This work introduces a data-driven model that can accurately and rapidly predict the power generation of all wind turbines in a given wind farm, regardless of the wind farm layout, yaw angle configurations and wind conditions. The proposed computational framework encodes the wind farm into a fully-connected graph and processes it through a graph transformer. The model was shown to generalise well, averaging a relative accuracy of $99.8\%$ over power prediction of previously unseen wind farm, while being able to identify latent relationships within wind farms that correspond well with the wind farm wake maps. The graph transformer model was used as a surrogate for PyWake in optimising yaw angle configurations of wind farms with genetic algorithms, and demonstrated exceptional accuracy in selecting the optimal yaw angle configurations, while simultaneously significantly reducing the computation time required. This encouraging result motivates further research into the potential of transfer learning graph transformers on wind farm simulations of higher fidelity or real-world wind farm data.

\section*{Acknowledgements}
AR was supported by an EPSRC CASE studentship supported by Shell, and AAF was supported by a UKRI Turing AI Fellowship (EP/V025449/1). 

\bibliographystyle{unsrt}
{\footnotesize\bibliography{citations.bib}}

\end{document}